# Output-Feedback Path Planning with Robustness to State-Dependent Errors

Mahroo Bahreinian[1], and Roberto Tron[2]

*Abstract*— We consider the problem of sample-based feedback motion planning from measurements affected by *systematic errors*. Our previous work presented output feedback controllers that use measurements from landmarks in the environment to navigate through a cell-decomposable environment using duality, Control Lyapunov and Barrier Functions (CLF, CBF), and Linear Programming. In this paper, we build on this previous work with a novel strategy that allows the use of measurements affected by systematic errors in perceived depth (similarly to what might be generated by vision-based sensors), as opposed to accurate displacements measurements. As a result, our new method has the advantage of providing more robust performance (with quantitative guarantees) when inaccurate sensors are used. We test the proposed algorithm in the simulation to evaluate the performance limits of our approach predicted by our theoretical derivations.

## I. INTRODUCTION

Motion planning is a major research area in the context of mobile robots, as it deals with the problem of finding a path from an initial state toward a goal state while avoiding collisions [5], [13], [14]. Traditional path planning methods focus on finding *single nominal paths* in a given *known map*, and the majority of them make the implicit assumption that the agent possesses a lower-level *state feedback* controller for following such nominal path in the face of external disturbances and imperfect models. In this paper, we instead use the alternative approach of synthesizing a set of output-feedback controllers instead of single a nominal path. By using this approach, we can directly synthesize path planning solutions that are robust to the inaccuracies and systematic errors that are present in real-world sensors, while also giving a formal characterization of the errors that can be tolerated. In particular, in this paper, we focus on errors in perceived depth when measuring landmarks in an environment using vision-based sensors.

*Previous works*. A well-known class of techniques for motion planning is sampling-based methods. Algorithms such as Probabilistic Road Maps (PRM) [11], Rapidly Exploring Random Trees (RRT*, [15], [17]) and asymptotically optimal Rapidly Exploring Random Tree (RRT*, [10]), have become popular in the last few years due to their good practical performance, and their probabilistic completeness [10], [16], [17]; there have also been extensions considering perception uncertainty [24]. However, these algorithms only provide nominal paths and assume that a separate low-level controller exists to generate collision-free trajectories at run time.

The concept of the output feedback controller is based on [2], [3], which designs a set of linear output feedback controllers that take as input relative displacement measurements with respect to a set of landmarks in different regions of the environment. This algorithm is based on solving a sequence of robust min-max Linear Programming (LP) problems in each region (which typically corresponds to a convex cell decomposition [2] or is derived from sampling [3]); the formulation of the optimization problem uses linear Control Lyapunov Function (CLF) and Control Barrier Function (CBF) constraints to provide stability and safety guarantees, respectively. A significant advantage of output feedback controllers is that they can automatically recover from deviations from a nominal path, and have empirically shown robustness to significant discrepancies between the actual environment and the map used for planning. In [26] the author uses the same method to compute a set of controllers which take the noisy measurements with known bounded uncertainties. The author considers Gaussian noise distributions centered on the landmark's true location while in this work, we consider uncertainties that can potentially depend on the position of the robot.

In order to observe the environment, modern robots can be equipped with different kinds of sensors, such as Inertial Measurement Units [29], visual sensors [22], sonar sensors [28], and contact sensors [19], etc. Among these, monocular cameras are nearly ubiquitous, given their favorable trade-offs with respect to size, power consumption, and the richness of information in the measurements; a peculiarity of this sensor is that, due to perspective projection, it can provide relatively precise *bearing* information (the direction of an object or point in the environment) but not, from a single image alone, *depth* information. A common problem with measurements obtained with monocular cameras is the fact that the estimated depths of the points might present large errors, due to the fact that depth can be accurately estimated only when large *baselines* are used, i.e., when the camera observes the same points from two or more relatively distant locations [7]. Even when depth cameras [8] is used, the error in the depth estimate is much more significant than the error in image coordinates, and its variance depends on the actual depth of the measurement as shown in Fig. 1 (i.e., further points have larger errors). Many different techniques have been introduced for monocular navigation and control. An

This work was supported by ONR MURI N00014-19-1-2571 "Neuro-Autonomy: Neuroscience-Inspired Perception, Navigation, and Spatial Awareness"

[1]Mahroo Bahreinian is with the Division of Systems Engineering at Boston University, Boston, MA, 02215 USA. Email: `mahroobh@bu.edu`

[2]Roberto Tron is with Faculty of Department of Mechanical Engineering and Systems Engineering at Boston University, Boston, MA, 02215 USA. Email: `tron@bu.edu`

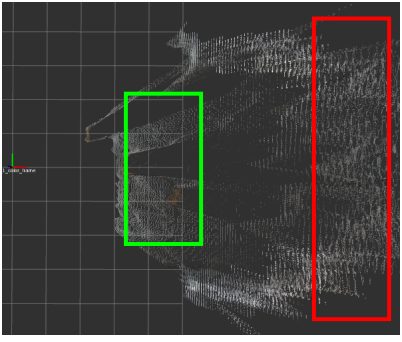

Fig. 1: Depth estimation from Realsense camera, points further from the camera (e.g., see red box) have larger absolute errors than points closer to the camera (e.g., see green box).

example is visual servoing, where image-based computations are used to find a control law that guides the robot toward a *home* location using bearing alone [1], [12], [18], [25], or by concurrently estimating depths [6], [20], [23].

*Proposed approach and contributions*. The goal of the present work is to build upon the aforementioned line of work: we apply the synthesis method of [2], [3] but with controllers that use as input a set of measurements between the robot and a set of landmarks. We assume that the relative displacement measurements are corrupted by bonded errors which are defined by the sensor characterization. Since we consider a planning problem, we assume that a map of the environment is available and that the positions of the landmarks are known and fixed in the reference frame of this map. One point to make is that, at a technical level, the problem we consider has state-dependent errors (points closer to the robot have smaller errors, in absolute terms; this gives a more complex structure to the set of constraints with respect to which we need to be robust.

The main contribution of this work is then to provide a way to synthesize feedback controllers that can be used to solve the path planning problem by extending the method of [2], [3] to take the measurements that are corrupted by systematic errors as an input. This contribution is evaluated via simulations. Our analysis of the results also shows a trade-off between guarantees on performance (stability) versus robustness to errors in the measurements.

## II. NOTATION AND PRELIMINARIES

In this section, we review the system dynamics, CLF and CBF constraints in the context of our proposed work. We use the same notation as [2].

### A. System Dynamics

We assume that the robot has linear dynamics of the form

$$\dot{x} = Ax + Bu, \quad (1)$$

where $x \in \mathcal{X} \subset \mathbb{R}^n$ denotes the state, $u \in \mathcal{U} \subset \mathbb{R}^m$ is the system input, and $A \in \mathbb{R}^{n \times n}$, $B \in \mathbb{R}^{n \times m}$ define the linear dynamics of the system. We assume that the pair $(A, B)$ is controllable, and that $\mathcal{X}_i$ and $\mathcal{U}$ are polytopic,

$$\mathcal{X} = \{x \mid A_x x \leq b_x\}, \quad \mathcal{U} = \{u \mid A_u u \leq b_u\}, \quad (2)$$

where $A_x \in \mathbb{R}^{p \times n}$, $b_x \in \mathbb{R}^p$, $A_u \in \mathbb{R}^{q \times m}$, $b_u \in \mathbb{R}^q$.

### B. Control Lyapunov and Barrier Functions (CLF, CBF)

In this section, we review the CLF and CBF constraints, which are differential inequalities that ensure stability and safety (set invariance) of a control signal $u$ with respect to the dynamics (1). First, it is necessary to review the following.

*Definition 1:* The Lie derivative of a differentiable function $h$ for the dynamics (1) with respect to the vector field $Ax$ is defined as $\mathcal{L}_{Ax} h(x) = \frac{\partial h(x(t))}{\partial x}^{\mathrm{T}} Ax$.

Applying this definition to (1) we obtain

$$\dot{h}(x) = \mathcal{L}_{Ax} h(x) + \mathcal{L}_B h(x) u. \quad (3)$$

In this work, we assume that Lie derivatives of $h(x)$ of the first order are sufficient [9] (i.e., $h(x)$ has relative degree 1 with respect to the dynamics (1)); however, the result could be extended to the higher degrees, as discussed in [2].

We now pass to the definition of the differential constraints. Consider a continuously differentiable function $V(x) : \mathcal{X} \to \mathbb{R}$, $V(x) \geq 0$ for all $x \in \mathcal{X}$, with $V(x) = 0$ for some $x \in \mathcal{X}$.

*Definition 2:* The function $V(x)$ is a *Control Lyapunov Function* (CLF) with respect to (1) if there exists positive constant $c_v$ and control inputs $u \in \mathcal{U}$ such that

$$\mathcal{L}_{Ax} V(x) + \mathcal{L}_B V(x) u + c_v^{\mathrm{T}} V(x) \leq 0, \forall x \in \mathcal{X}. \quad (4)$$

Furthermore, (4) implies that $\lim_{t \to \infty} V(x(t)) = 0$.
Note that, in (4), $c_v$ denotes as a decay rate of $V(x)$, so if $c_v = 0$, it implies that the system (1) is stable but not exponentially stable while a large $c_v$ means that the system (1) is exponentially stable and the robot is required to make rapid progress toward the goal.

Now consider a continuously differentiable function $h(x) : \mathcal{X} \to \mathbb{R}$ which defines a safe set $\mathcal{C}$ such that

$$\begin{aligned} \mathcal{C} &= \{x \in \mathbb{R}^n \mid h(x) \geq 0\}, \\ \partial \mathcal{C} &= \{x \in \mathbb{R}^n \mid h(x) = 0\}, \\ Int(\mathcal{C}) &= \{x \in \mathbb{R}^n \mid h(x) > 0\}. \end{aligned} \quad (5)$$

In our setting, the set $\mathcal{C}$ will represent a convex local approximation of the free configuration space (in the sense that $x \in \mathcal{C}$ does not contain any sample that was found to be in a collision). We say that the set $\mathcal{C}$ is *forward invariant* (also said *positive invariant* [4]) if $x(t_0) \in \mathcal{C}$ implies $x(t) \in \mathcal{C}$, for all $t \geq 0$ [27].

*Definition 3 (CBF, [21]):* The function $h(x)$ is a *Control Barrier Function* with respect to (1) if there exists a positive constant $c_h$, control inputs $u \in \mathcal{U}$, and a set $\mathcal{C}$ such that

$$\mathcal{L}_{Ax} h(x) + \mathcal{L}_B h(x) u + c_h^{\mathrm{T}} h(x) \geq 0, \forall x \in \mathcal{C}. \quad (6)$$

Furthermore, (6) implies that the set $\mathcal{C}$ is forward invariant. Note that in (6), $c_h$ denotes as a maximum decay rate of $h(x)$. If $c_h = 0$, then the robot is not allowed to move toward the obstacle while a larger $c_h$ means that the robot is allowed to move toward the obstacle very fast.

## C. Stability via a Control Lyapunov Function

For each cell $\mathcal{X}$ in the decomposition of the environment, we define an *exit face* to be the face which the robot passes through to reach the next region. To stabilize the navigation along the exit direction of the convex cell $\mathcal{X}$, we define the Lyapunov function $V(x)$ as

$$V(x) = z^T(x - x_e), \tag{7}$$

where $z$ is the unit vector of the exit direction for the convex cell $\mathcal{X}$, $x_e \in$ *exit face* is a point belongs to the exit face, and $V(x)$ reaches its minimum $V(x) = 0$ at $x_e$. Note that the Lyapunov function represents, the distance $d(x, x_e)$ between the current system position and the exit face. By Definition 2, $V(x)$ is a CLF.

## D. Safety via Control Barrier Functions

Let $A_h, b_h$ be obtained from $A_x, b_x$ by eliminating the row corresponding to the exit face. The corresponding CBF is then defined as

$$h(x) = A_h x + b_h. \tag{8}$$

## III. FEEDBACK CONTROL PLANNING

As mentioned in the introduction, in this paper we assume that the robot is equipped with an inaccurate sensor that measures the relative displacement between its position and a set of landmarks with bounded errors in the depth (in this paper, we assume that the bearing error is negligible with respect to the depth error). In this section, we discuss how to find a linear controller that, given the noisy displacement measurements, provides a control command to drive the robot toward the exit direction while avoiding the walls of the cell. We divide this section into three parts. First, we give details on our model for the erroneous measurements and the controller; second, we discuss our approach to find a control gain matrix that is robust to the errors in the measurements by applying duality theory twice. Then, we discuss a bisection search on the error bounds to (intuitively) maximize the robustness of the controller.

## A. Measurement Model and Output Feedback Controller

We assume that the robot can only measure the relative displacements between the robot's position $x$ and the landmarks in the environment with errors in their depth; more precisely, this is modeled using the output function

$$\mathcal{Y} = \left(\mathrm{diag}(s)(L - x\mathbf{1}^{\mathrm{T}})\right)^{\vee} \tag{9}$$

where $L \in \mathbb{R}^{n \times N}$ is a matrix of landmark locations and $N$ is the number of landmarks, $A^{\vee}$ represents the vectorized version of a matrix $A$ and $s \in \mathbb{R}^N$ is a vector that contains proportional scaling errors multiplying each measurement. We assume that the error vector $s$ is bounded by

$$s_{\min} \leq s \leq s_{\max}, \tag{10}$$

where $s_{\min}$ and $s_{\max}$ are vectors; we discuss the choice of $s_{\min}$ and $s_{\max}$ in Section III.C. Note that the case $s = \mathbf{1}_{nN}$ corresponds to the case where all the measurements are accurate.

*Remark 1:* Our model has the property that the absolute error between measured and actual displacements depends on the magnitude of the actual displacement (i.e., points that are further away have larger absolute errors); this reflects the behavior of real sensors (see Fig. 1).

We define the feedback controller as

$$u = \mathcal{K}\mathcal{Y} + k, \tag{11}$$

where $\mathcal{K} \in \mathbb{R}^{m \times nN}$ and $k \in \mathbb{R}^m$ are the feedback gains that need to be found for the convex cell $\mathcal{X}$.

## B. Computation of the Controller via Linear Programming

The input to the controller $u$ is the imprecise relative displacement measurements $\mathcal{Y}$ between the position of the robot and the set of landmarks in the environment. The controller in (11) is a weighted linear combination of the inputs, given by the matrix $K$, plus a vector $k$. The goal is to design $K$ and $k$ such that the system is driven toward the exit direction $z$ while avoiding the obstacles (walls). Assume the controller is in the form of (11), and the output function is (9), following the approach of [2], [3], and using the CLF-CBF constraints reviewed in Sec. II, we encode our goal in the following feasibility problem:

$$\begin{aligned}
&\text{find } \{\mathcal{K}, k\} \\
&\text{subject to:} \\
&\left\{\text{CBF:} - (\mathcal{L}_{Ax}h(x) + \mathcal{L}_B h(x)u + c_h^{\mathrm{T}}h(x))\right\} \leq 0, \\
&\left\{\text{CLF:} \quad \mathcal{L}_{Ax}V(x) + \mathcal{L}_B V(x)u + c_v^{\mathrm{T}}V(x)\right\} \leq 0, \\
&u \in \mathcal{U}, \ \forall x \in \mathcal{X}, \ \forall s \in (s_{\min}, s_{\max}).
\end{aligned} \tag{12}$$

The constraints in (12) need to be satisfied for all $x$ in the region $\mathcal{X}$, i.e., the same control gains should satisfy the CLF-CBF constraints at every point in the region and for all possible realization of the errors $s$ in the bounds given by $(s_{\min}, s_{\max})$. Following [3], the problem (12) is rewritten using a min-max formulation:

$$\begin{aligned}
&\min_{\mathcal{K}, k, \delta_v, \delta_h} \ w_h^{\mathrm{T}}\delta_h + w_v \delta_v \\
&\text{subject to} \ \max_{x \in \mathcal{X}} \text{CBF} \leq \delta_h, \\
&\qquad\qquad \max_{x \in \mathcal{X}} \text{CLF} \leq \delta_v, \\
&\delta_v, \delta_h \leq 0, \ \forall s \in (s_{\min}, s_{\max}),
\end{aligned} \tag{13}$$

where $\delta_h, \delta_v$ are slack variables, and the weights $w_h$ and $w_v$ are user-defined constants defining the trade-off between the barrier functions and Lyapunov function constraints. From (3), the Lie derivatives of $h(x)$ and $V(x)$ are written as:

$$\begin{aligned}
\dot{h}(x) &= \mathcal{L}_{Ax}h(x) + \mathcal{L}_B h(x)u = A_h(Ax + Bu), \\
\dot{V}(x) &= \mathcal{L}_{Ax}V(x) + \mathcal{L}_B V(x)u = z^{\mathrm{T}}(Ax + Bu).
\end{aligned} \tag{14}$$

Substituting the measurement model (9) and the controller (11) in the Lie derivatives in (14) for the dynamics (1), the

constraints in (13) can be rewritten as:

CBF constraint:
$$\begin{bmatrix} \max_x & -(A_h A - A_h B \mathcal{K} \operatorname{diag}(s)\mathcal{I} + c_h A_h)x \\ & \text{subject to} \quad A_x x \leq b_x \end{bmatrix} \leq \\ \delta_h + c_h b_h + A_h B \mathcal{K} \operatorname{diag}(s) L^\vee + A_h k, \quad (15)$$

CLF constraint:
$$\begin{bmatrix} \max_x & z^T A - z^T B \mathcal{K} \operatorname{diag}(s)\mathcal{I} + c_v z^T x \\ & \text{subject to} \quad A_x x \leq b_x \end{bmatrix} \leq \\ \delta_v + c_v z^T x_e - z^T B \mathcal{K} \operatorname{diag}(s) L^\vee - z^T k, \quad (16)$$

where $\mathcal{I} = \mathbf{1}_N \otimes I_n$. At first sight, this min-max optimization problem might appear very challenging, as the constraints in (15), (16) are trilinear in terms of the variables $K$, $x$ and $s$. However, we show below that we can reduce the problem to a simple Linear Program by repeated application of duality. We first write the dual forms of the constraints with respect to $x$:

CBF dual constraint:
$$\begin{bmatrix} \min_{\lambda_h} b_x^T \lambda_h \\ \text{subject to} \\ A_x^T \lambda_h \geq (-A_h A + A_h B \mathcal{K} \operatorname{diag}(s)\mathcal{I} - c_h A_h)^T \\ \lambda_h \geq 0, \end{bmatrix} \leq \\ \delta_h + c_h b_h + A_h B \mathcal{K} \operatorname{diag}(s) L^\vee + A_h k, \quad (17)$$

CLF dual constraint:
$$\begin{bmatrix} \min_{\lambda_l} b_x^T \lambda_l \\ \text{subject to} \\ A_x^T \lambda_l \geq (z^T A - z^T B \mathcal{K} \operatorname{diag}(s)\mathcal{I} + c_v z^T)^T \\ \lambda_l \geq 0 \end{bmatrix} \leq \\ \delta_v + c_v z^T x_e - z^T B \mathcal{K} \operatorname{diag}(s) L^\vee - z^T k, \quad (18)$$

Consequently, (13) with the dual constraints becomes:
$$\min_{\mathcal{K}, k, \delta_v, \delta_h} w_h^T \delta_h + w_v \delta_v$$
subject to:
$$b_x^T \lambda_h \leq \delta_h + c_h b_h + A_h B \mathcal{K} \operatorname{diag}(s) L^\vee + A_h k, \quad (19a)$$
$$-A_x^T \lambda_h \leq (A_h A - A_h B \mathcal{K} \operatorname{diag}(s)\mathcal{I} + c_h A_h)^T, \quad (19b)$$
$$b^T \lambda_l \leq \delta_v + c_v z^T x_e - z^T B \mathcal{K} \operatorname{diag}(s) L^\vee - z^T k, \quad (19c)$$
$$-A_x^T \lambda_l \leq (-z^T A + z^T B \mathcal{K} \operatorname{diag}(s)\mathcal{I} - c_v z^T)^T, \quad (19d)$$
$$\delta_h \leq 0, \quad \delta_v \leq 0, \lambda_h \geq 0, \lambda_l \geq 0, \forall s \in (s_{\min}, s_{\max}).$$

From [2, Lemma 1, Lemma 2], the optimization problem (13) is equivalent to (19).

Up to this point, the optimization problem (19) represents a small modification to the results from [2]; however, the key insight is that all the constraints in (19) are *linear* in the unknown $s$. Since we know the constraints on the error vector $s$, we can find the controller that is feasible for all $s \in (s_{\min}, s_{\max})$ by applying again a similar procedure: we rewrite (19) in terms of the min-max optimization, so we convert each constraints in (19) to a maximization problem as follows,

constraint (19a):
$$\begin{bmatrix} \max_s -A_h B \mathcal{K} \operatorname{diag}(s) L^\vee \\ \text{subject to} \\ s_{\min} \leq s \leq s_{\max} \end{bmatrix} \leq \delta_h + c_h b_h + A_h k - b_x^T \lambda_h. \quad (20)$$

The maximization problem in (20) is linear in terms of variable $s$, so we can write dual forms of (20) and convert the minimization problem (see Appendix (24)) as,
$$\begin{bmatrix} \min_{p_1} \mathcal{S}_b^T p_1 \\ \text{subject to} \\ \mathcal{S}_A^T p_1 \geq M_1, \; p_1 \geq 0 \end{bmatrix} \leq \delta_h + c_h b_h + A_h k - b_x^T \lambda_h. \quad (21)$$

where $\mathcal{S}_A$ and $\mathcal{S}_b$ are defined in (23) and $M_1$ is defined in (24) in the Appendix. Similar to (21), we can rewrite constraints (19b), (19c), and (19d) in terms of the minimization problem and convert (19) to a minimization problem over all $s \in (s_{\min}, s_{\max})$, which is explained in details in the Appendix.

*Proposition 1:* Minimizing the cost function in (19) subject to constraints in form of (21) has the same optimal feasible solution as the optimization problem (13).

*Proof:* From [2, Lemma 1, Lemma 2], minimizing cost function in (19) subject to constraints in form of (21) is equivalent to solve (19), and optimization problem (19) is equivalent to (13), so minimizing cost function in (19) subject to constraints in form of (21) has the same feasible optimization solution as (13). ∎

### C. Discussion on Error Bounds

We can rewrite the controller in (11) as,
$$u = \mathcal{K} \operatorname{diag}(s_{\min}) \operatorname{diag}(s_{\min})^{-1} \operatorname{diag}(s)\mathcal{Y} + k, \quad (22)$$

and we define $\mathcal{K}' = \mathcal{K} \operatorname{diag}(s_{\min})$ and $s' = \operatorname{diag}(s_{\min})^{-1} s$, the constraint in (20) becomes $1 \leq s' \leq \operatorname{diag}(s_{\min})^{-1} s_{\max}$. We denote $s'_{\max} = \operatorname{diag}(s_{\min})^{-1} s_{\max}$.

In order to compute the control gain matrix, we perform a bisection search to find the largest $s'_{\max}$ the optimization problem has feasible solution. As shown in Fig. 4, as $c_v$ decreases the optimization problem can tolerate larger value for $s'_{max}$. Note that we can rewrite the controller in (22) in terms of $s_{\min}$ and choose $s'_{\min}$ instead.

## IV. NUMERICAL EXAMPLE

In this section, we apply our proposed method to a cell decomposition of the configuration space in Fig. 2a. While the optimization problem guarantees safety and stability constraints, in these experiments the velocity control input $u$ has been normalized to achieve constant velocities along with cells.

For each cell in Fig. 2a, we compute the controller such that it moves the robot toward the exit direction while avoiding colliding with the boundaries of the cell. Given the set of computed control gains, we test the controller for different values of $s$. We assume the robot starts from location $[10; 30]$ and $s \in (\frac{1}{1.5}, 1.5)$. In Fig. 2b, we choose $s = 1$ which

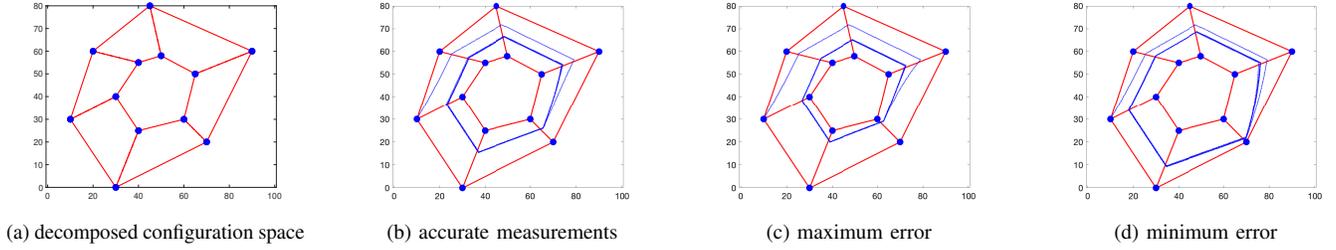

(a) decomposed configuration space  (b) accurate measurements  (c) maximum error  (d) minimum error

Fig. 2: In Fig. 2a, the configuration space is decomposed into 6 convex cells. The blue dots are the landmarks in the environment, the dashed red lines are the exit faces of the convex cell that the robot passes through

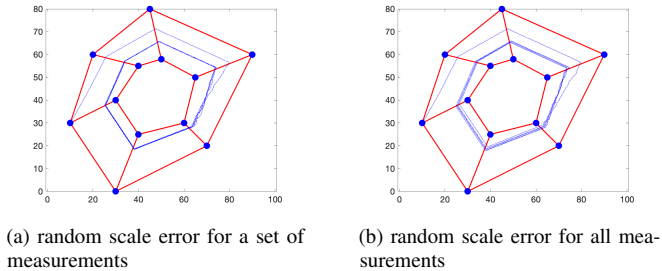

(a) random scale error for a set of measurements

(b) random scale error for all measurements

Fig. 3: In Fig. 3a, a set of measurements are chosen randomly and scaled with random bounded errors while in Fig. 3b, all measurements are scaled with random bounded errors.

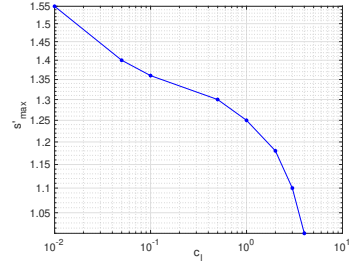

Fig. 4: Effect of $c_v$ on $s'_{\max}$ with constant $c_h = 1$

means all the relative displacement measurements are accurate. Then, in Fig. 2c, we pick $s = 1.5$ which means all the relative displacement measurements are scaled by error 1.5, and in Fig. 2d, we pick $s = \frac{1}{1.5}$ which means all the relative displacement measurements are multiplied by a coefficient $\frac{1}{1.5}$. In all these examples, the robot patrols the environment without violating safety and stability constraints. In Fig. 3b, at each time step, we randomly select two landmarks and multiply the relative displacement measurements assigned to those landmarks by random scale errors between $\frac{1}{1.5}$ and 1.5. As shown in Fig. 3a the robot patrols cells when a portion of measurements are corrupted by random scale errors. Then, in Fig. 3b the robot patrols cells when all measurements are corrupted by random scale errors. Comparing Fig. 3a to Fig. 3b, the deviates from the path in Fig. 2b is more in Fig. 3b rather than Fig. 3a.

### A. Considerations on the choice of parameters

In this section, we discuss numerical analysis such as dependencies of $s'_{\max}$ on parameters of the problem, $c_v$ and $c_h$. As mentioned in Sec. III-C, we perform the bisection search to achieve the largest $s'_{\max}$. We start our search by changing the value of $c_v$ and study the effect of this change on the constraint on $s$. The result of this search is shown in Fig. 4, which implies by decreasing $c_v$, we can reach larger $s'_{\max}$ such that the solution of optimization problem (12) is feasible. By decreasing $c_v$, we sacrifice the exponential stability of the system to gain robustness to larger scale errors. Then, we implement the same experiment on $c_h$, however, chaining $c_h$ does not affect the constraint on $s$ and we plan to analyze and explain it in our future work.

To solve the optimization problem (12) we also introduce constraints to bound $k$. Without this constraint on $k$, we might find matrix $\mathcal{K}$ close to zero which eliminates the fact that the controller is the linear combination of the displacement measurements.

## V. CONCLUSIONS

In this paper we proposed a novel approach to design a output-feedback controller on a cell decomposition, through Linear Programming. The input to the controller is the relative displacement measurements with respect to the landmarks which are corrupted by systematic errors. To compute the control gains, we formed the min-max convex problem and then we changed the min-max optimization problem to min-min optimization problem by forming the dual of the inner maximization problems while also giving a formal characterization of the errors that can be tolerated. Our out-put feedback controller is robust to the inaccuracies and systematic errors that are present in real-world sensors, in particular, we focus on errors in perceived depth when measuring landmarks in an environment using vision-based sensors. We test the proposed algorithm in the simulation to evaluate the performance limits of our approach predicted by our theoretical derivations.

## APPENDIX

We can rewrite the constraint $s \in (s_{\min}, s_{\max})$ as

$$\begin{bmatrix} I_N \\ -I_N \end{bmatrix} s \leq \begin{bmatrix} s_{\max} \\ -s_{\min} \end{bmatrix}, \quad S_A = \begin{bmatrix} I_N \\ -I_N \end{bmatrix}, \quad S_b = \begin{bmatrix} s_{\max} \\ -s_{\min} \end{bmatrix} \tag{23}$$

Define $\mathcal{I}_s = 1_{nl}^T \otimes I_2$ and $\mathbf{1} = I_N \otimes [1,1]$, where $N$ is the number of landmarks, then

$$\begin{aligned}
-A_h B\mathcal{K} \operatorname{diag}(s) L^\vee &= -s^T (\operatorname{diag}(L^\vee A_h B\mathcal{K})^T \mathbf{1}^T)^T = s^T M_1 \\
[A_h \mathcal{K} \operatorname{diag}(s)\mathcal{I}]_i &= s^T (\operatorname{diag}([\mathcal{I}_s]_i A_h B\mathcal{K})^T \mathbf{1}^T)^T = s^T M_2^i \\
-z^T \mathcal{K} \operatorname{diag}(s) L^\vee &= -s^T (\operatorname{diag}(L^\vee z^T \mathcal{K})^T \mathbf{1}^T)^T = s^T M_3 \\
[z^T \mathcal{K} \operatorname{diag}(s)\mathcal{I}]_i &= -s^T (\operatorname{diag}([\mathcal{I}_s]_i z^T \mathcal{K})^T \mathbf{1}^T)^T = s^T M_4^i.
\end{aligned} \quad (24)$$

for $i \in \{1,\ldots,n\}$ and $[H]_i$ denotes as the $i^{th}$ row of matrix $H$. We use (24) to rewrite constraints (19a)-(19d) in form of maximization problem as:

constraint (19b)
$$\begin{bmatrix} \max_s s^T M_2 \\ \text{s.t} \\ \mathcal{S}_A s \le \mathcal{S}_b \end{bmatrix} \le (A_h A + c_h A_h)^\mathrm{T} + A_x^\mathrm{T} \lambda_h \quad (25)$$

constraint (19c)
$$\begin{bmatrix} \max_s s^T M_3 \\ \text{s.t} \\ \mathcal{S}_A s \le \mathcal{S}_b \end{bmatrix} \le \delta_v + c_v z^\mathrm{T} x_e - z^\mathrm{T} k - b_x^\mathrm{T} \lambda_l, \quad (26)$$

constraint (19d)
$$\begin{bmatrix} \max_s s^T M_4 \\ \text{s.t} \\ \mathcal{S}_A s \le \mathcal{S}_b \end{bmatrix} \le -(z^\mathrm{T} A + c_v z^\mathrm{T})^\mathrm{T} + A_x^\mathrm{T} \lambda_l \quad (27)$$

We form the dual of the maximization problems (25)-(27), and rewrite

$$\begin{aligned}
\min_{\delta_h, \delta_v, \mathcal{K}, k} \quad & w_h^\mathrm{T} \delta_h + w_v \delta_v \\
\text{s.t:} \quad & \mathcal{S}_b^T P_1 \le \delta_h + c_h b_h + A_h k - b^T \lambda_h, \\
& \mathcal{S}_A^T P_1 \le M_1 \\
& \mathcal{S}_b^T P_2 \le c_h A_h^T + A^T \lambda_h \\
& \mathcal{S}_A^T P_2^i \le M_2^i \\
& \mathcal{S}_b^T P_3 \le \delta_v + c_v z^T x_e - z^T k - b^T \lambda_l, \\
& \mathcal{S}_A^T P_3 \le M_3 \\
& \mathcal{S}_b^T P_4 \le -c_v z^T + A^T \lambda_l \\
& \mathcal{S}_A^T P_4^i \le M_4^i \\
& \delta_v \le 0, \delta_h \le 0 \\
& \lambda_h \ge 0, \lambda_l \ge 0 \\
& P_1 \ge 0, P_2^i \ge 0, P_3 \ge 0, P_4^i \ge 0,
\end{aligned} \quad (28)$$

for $i \in \{1,\ldots,n\}$.